\documentclass{bmvc2k}

\usepackage{amsmath}
\usepackage{color}
\usepackage{algorithm2e}
\usepackage[noend]{algpseudocode}
\usepackage{varwidth}
\usepackage{graphicx}
\usepackage{caption}
\usepackage{hyperref}
\usepackage[table]{xcolor}
\usepackage{tabularx}
\usepackage{floatrow}

\newfloatcommand{capbtabbox}{table}[][\FBwidth]
\newcommand{\argmin}{\mathop{\mathrm{argmin}}}
\newcommand{\nosemic}{\renewcommand{\@endalgocfline}{\relax}}% Drop semi-colon ;
\newcommand{\dosemic}{\renewcommand{\@endalgocfline}{\algocf@endline}}

\title{Understanding Deep Architectures by Visual Summaries}

\addauthor{Marco Godi}{marco.godi@univr.it}{1}
\addauthor{Marco Carletti}{marco.carletti@univr.it}{1}
\addauthor{Maedeh Aghaei}{aghaei.maya@gmail.com}{2}
\addauthor{Francesco Giuliari}{francesco.giuliari@studenti.univr.it}{1}
\addauthor{Marco Cristani}{marco.cristani@univr.it}{1}

\addinstitution{
 University of Verona\\
 Italy
}
\addinstitution{
 University of Barcelona\\
 Spain
}

\runninghead{Godi, Carletti}{Understanding Deep Architectures by Visual Summaries}

\begin{document}

\maketitle \vspace{0cm}

\begin{abstract}
In deep learning, visualization techniques extract the salient patterns exploited by deep networks for image classification, focusing on single images; no effort has been spent in investigating whether these patterns are systematically related to precise semantic entities over multiple images belonging to a same class, thus failing to capture the very understanding of the image class the network has realized. This paper goes in this direction, presenting a visualization framework which produces a group of clusters or \emph{summaries}, each one formed by crisp salient image regions focusing on a particular part that the network has exploited with high regularity to decide for a  given class. The approach is based on a sparse optimization step providing sharp image saliency masks that are clustered together by means of a semantic flow similarity measure. The summaries communicate clearly what a network has exploited of a particular image class, and this is proved through automatic image tagging and with a user study. Beyond the deep network understanding, summaries are also useful for many quantitative reasons: their number is correlated with ability of a network to classify (more summaries, better performances), and they can be used to improve the classification accuracy of a network through summary-driven specializations. 
\end{abstract}

%-------------------------------------------------------------------------
\section{Introduction}
\label{sec:intro}
Individuating the visual regions exploited by a deep network for making decisions is important: this allows to 
foresee potential failures and highlight differences among diverse network architectures \cite{zhou2014object,yosinski2015understanding,zintgraf2017visualizing,zeiler2014visualizing}.
This is the goal of the \emph{visualization} strategies: %, that the researcher has to interpret for the analysis.
early work \cite{zhou2014object,deconvnet, dosovitskiy2016inverting,yosinski2015understanding} individuate those images which activate a certain neuron the most; other approaches consider the network as a whole, generating dreamlike images bringing the classifier to high classification scores ~\cite{simonyan2013deep,zeiler2014visualizing,olah2018the}. The most studied type of visualization techniques however, highlights those salient patterns which drive a classifier toward a class ~\cite{erhan2009visualizing,Fong_2017_ICCV,gradcam,cam, guidedbackprop_vedaldi,zhang2016top} or against it~\cite{zintgraf2017visualizing} through smooth saliency maps.

However, no prior study investigated whether these salient patterns are systematically related to precise semantic entities to describe an object class. In fact, the previous visualization systems analyze single images independently, and no reasoning on multiple images from the same class is carried out. In other words, these approaches are not able to reveal if a network has captured an object class in all of its local aspects. It would be of great importance for interpretation of deep-architectures to be able to understand for example, that AlexNet when classifying the class "golden retriever" is systematically very sensible to the visual patterns representing the nose, the eye and the mouth, so that the absence of one or all of these patterns in an image will most probably bring to a failure. At the same time, knowing that GoogleNet has understood also the tail (in addition to the previous parts) can add a semantic explanation of its superiority w.r.t. AlexNet. 

\begin{figure*}[t!]
   \includegraphics[width=1.0\linewidth]{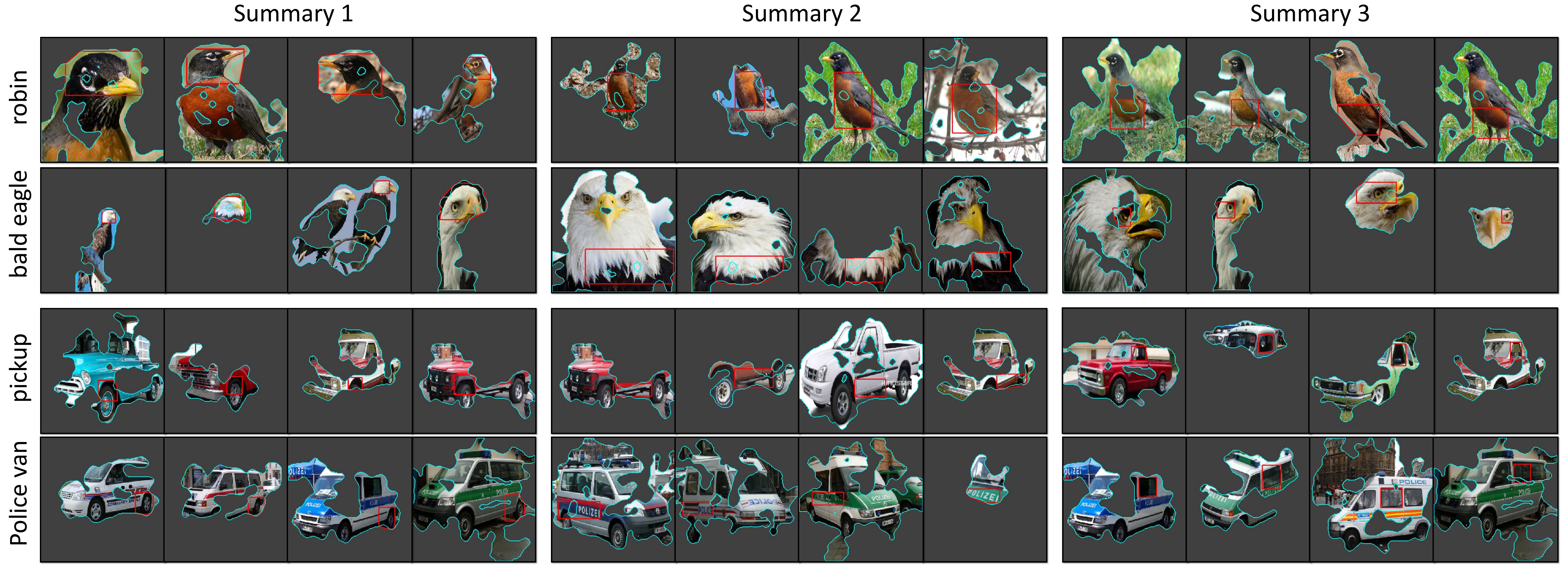}  \vspace{-0.6cm}
    \captionof{figure}{Visual summaries for AlexNet \cite{krizhevsky2012imagenet}. Each summary contains crisp salient regions, where common semantic parts are highlighted in red. It is easy to see that, i.e. in the \emph{robin} class, the network systematically considers the head (Summary 1), the body (Summary 2), the legs and the lower body (Summary 3). Best seen in color. \label{fig:figtitle}}
\end{figure*}

%are successful in \emph{conveying clear messages to the researchers}, fostering human explanations that are expressive (much information is disclosed)  and highly shared (diverse researchers give the same explanation). 
% In this regard, a visual analytics approach which provides interpretable visualizations has to guarantee that the explanations given to researchers about a classifier are expressive (they communicate some information) and highly shared (diverse researchers interpret the visualization in the same fashion); no user studies of this kind have been collected so far. 

% In this work, we design the first visualization approach which considers multiple images at a time, explaining what has been understood by a network in term of visual parts forming an object class. In practice, our approach takes as input a trained deep network and a test set, and provides as output a set of image clusters, or \emph{summaries}, where each cluster is representative of an object part. 

In this work, we present the first visualization approach which employs analysis of multiple images within an object class to provide an explanation on what has been understood by a network in terms of \emph{visual parts} to form an object class. In practice, our approach takes as input a trained deep network and a set of images, and provides as output a set of image clusters, or \emph{summaries}, where each cluster is representative of an object visual part.

% \begin{itemize}
% \item an improvement over a recent visualization technique, providing sharp regions the classifier focus on for the classification;
% \item a visual understanding system indicating the visual parts a classifier consistently uses for determining a class;
% \item a user study validating the usability of the system, showing that the users quickly understand these visual parts with high inter-rater reliability.
% \end{itemize}

Our visualization approach is composed by two phases. In the first phase, a crisp image saliency map is extracted from each test image, indicating the most important visual patterns for a given class. Important visual patterns are those that if perturbed in an image, lead to a high classification loss. The perturbation masks are found by an optimization process borrowed from \cite{Fong_2017_ICCV} and made sparse to provide binary values which results to a so called crisp mask. In facts, most literature on visualization provide smooth masks where higher values mean higher importance in the region \cite{zhou2014object, zintgraf2017visualizing, dosovitskiy2016inverting, Fong_2017_ICCV, gradcam, cam, guidedbackprop_vedaldi,zhang2016top}.
In this work however, we empirically demonstrate that our proposed crisp mask brings to higher classification loss w.r.t. smooth mask by incorporating a model to remove noisy patterns. Crisp mask on the other hand, facilitates further computations in the formation of the summaries.

In the second phase, the connected components, i.e. \emph{regions}, of the crisp masks are grouped across the image employing the affinity propagation algorithm ~\cite{affinitypropagation2007}, where the similarity measure is borrowed from the proposal flow algorithm \cite{proposalflow2016}. This allows for example to cluster together the wheel regions of different images from the car class, which together with other region clusters, facilitate interpretation of the class.
% the final visual explanations showed to the researchers, in the form of images exhibiting the crisp masks associated to the cluster.  

In the experiments, we show that our summaries capture clear visual semantics of an object class, by means of an automatic tagger and a user study. In addition, we show that the number of summaries produced by our approach is correlated with the classification accuracy of a deep network: the more the summaries, the higher the classification accuracy as demonstrated for AlexNet, VGG, GoogleNet, and ResNet in our experiments. Finally, we demonstrate that the summaries may improve the classification ability of a network, by adopting multiple, specific specialization procedures with the images of each summary.

The main contributions of this paper are as follows:

\begin{itemize}
\item Introduction of the first deep network saliency visualization approach to offer an understanding of the visual parts of an object class which are used for classification. 
\item Proposal of a model for crisp saliency mask extraction built upon the proposed model by \cite{Fong_2017_ICCV}.
\item Generation of visual summaries by grouping together crisp salient regions of commonly repetitive salient visual parts among multiple images within a same object class.
\item Presentation a comprehensive quantitative, qualitative, and human-based evaluation measures to demonstrate the advantages of visual summaries in terms of interpretability and possible applications.
\end{itemize}

%is the first network saliency visualization approach for multiple images at the same time. We also propose an improvement on the method for the saliency mask extraction approach described by \cite{Fong_2017_ICCV}, making it possible to generate masks with crisp boundaries. We show the advantages of this approach in terms of interpretability and possible applications.

%The rest of the paper is organized as follows: Sec.\ref{related_works} reviews the related works; our system is analyzed in Sec.\ref{method}, with the experiments following in Sec.\ref{experiments}. Finally, Sec.\ref{conclusions} concludes the paper with some observations and future perspectives.

\section{Related Work}
\label{related_works}

Visualization approaches can be categorized mainly into the \emph{local} and \emph{global} techniques. Local techniques focus on the understanding of single neurons by showing the filters or the activations \cite{yosinski2015understanding}. Under this umbrella, \emph{input-dependent} approaches select the images which activate a neuron the most \cite{zhou2014object,zeiler2014visualizing,dosovitskiy2016inverting}. \emph{Global} approaches however, capture some general property of the network, as like the tendency in focusing on some parts of the images for the classification \cite{zintgraf2017visualizing, simonyan2013deep, Fong_2017_ICCV, gradcam, cam, mahendran2015understanding}. These approaches are given a single image as input, and output a smooth saliency map in which the areas important for classification into a certain class are highlighted. Global approaches are mostly \emph{gradient-based}, computing the gradient of the class score with respect to the input image \cite{zeiler2014visualizing, dosovitskiy2016inverting, gradcam, mahendran2015understanding}. Our approach fall into the global category. Some other types of gradient-based approaches adds activations to the analysis, obtaining edge-based images with edges highlighted in correspondence of salient parts
% in which the edges are more intense where the saliency maps are more intense 
~\cite{gradcam}. Notably,  the technique of~\cite{zintgraf2017visualizing} individuates also the pixels which are \emph{against} a certain class. \emph{Generative} approaches generate dreamlike images bringing the classifier to high classification scores \cite{simonyan2013deep,nguyen2015deep,olah2018the}. In particular, the work of~\cite{olah2018the} is heavily built on generative-based \emph{local} representations, which are   
somewhat difficult to interpret, making the forecasting of the performance of the network against new data particularly complicated. \emph{Perturbation-based} approaches edit an input image and observe its effect on the output \cite{zintgraf2017visualizing}. In this case, the general output of the model is a saliency map showing how crucial is the covering of a particular area, that can be a pixel~\cite{zhou2014object, Fong_2017_ICCV} or superpixel-level map~\cite{lime}. In all of the previous cases, the outputs are single masked images. Our approach is also perturbation based, since it looks for crisp portions of images that if perturbed, maximally distract the classifier. However, unlike aforementioned models where the user has to interpret multiple saliency maps to explain the behavior of a particular classifier on a particular class, our proposed approach by providing visual summaries from the saliency maps, facilitates the interpretation task for the user.

\begin{figure*}
   \includegraphics[width=1.0\linewidth]{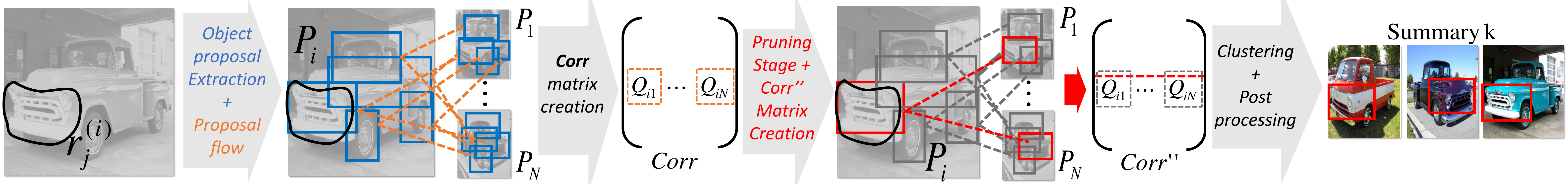} \vspace{-0.5cm}
    \captionof{figure}{Sketch of the clustering phase of our proposed method (Sec.~\ref{method_visualsummarization}). The pipeline starts with region proposal computation and Proposal Flow-based matching. The region proposals are pruned using overlap measurement on the saliency maps. The resulting matrix of compatibility values is then used as input for a clustering algorithm.}
    \label{fig:figmethodcluster} 
\end{figure*}

\section{Method}
\label{method}

% \begin{algorithm}
% \caption{My algorithm}\label{sharpmask}
% \begin{algorithmic}[1]
% \Procedure{MyProcedure}{}
% \State $\textit{stringlen} \gets \text{length of }\textit{string}$
% \State $i \gets \textit{patlen}$
% \If {$i > \textit{stringlen}$} \Return false
% \EndIf
% \State $j \gets \textit{patlen}$
% \If {$\textit{string}(i) = \textit{path}(j)$}
% \State $j \gets j-1$.
% \State $i \gets i-1$.
% \State \textbf{goto} \emph{loop}.
% \State \textbf{close};
% \EndIf
% \State $i \gets i+\max(\textit{delta}_1(\textit{string}(i)),\textit{delta}_2(j))$.
% \State \textbf{goto} \emph{top}.
% \EndProcedure
% \end{algorithmic}
% \end{algorithm}

Our method is composed by two phases, \emph{mask extraction} and \emph{clustering}. The former captures what visual patterns are maximally important for the classifier, and the latter organizes the visual patterns into summaries.  

\subsection{Mask Extraction}
\label{method_saliency}

Let us define a classifier as a function $y = f(x)$ where $x$ is the input image and $y$ is the classification score vector, in our case the softmax output of the last layer of a deep network. generating an output image in a global fashion.\\
Our starting point is the gradient-based optimization of~\cite{Fong_2017_ICCV}. In that method, the output of the optimization is a mask $m: \Lambda \rightarrow [0, 1]$ with the same resolution of $x$, in which higher values mean higher saliency.
The original optimization equation (Eq. (3) of~\cite{Fong_2017_ICCV}) is
\begin{equation}
\label{eq:orig}
m = \argmin_{m \in \left[ 0, 1 \right]^\Lambda} f_c(\Phi(x;m) ) + \lambda_1 \left\lVert 1 - m \right\rVert_1
\end{equation}
$\!\!$where $\Phi(x;m)$ is a perturbed version of $x$ in correspondence of the non-zero pixels of $m$, in which the perturbation function $\Phi$ does blurring: $
%\begin{equation}
\left[\Phi(x;m) \right](u)=\smallint g_{\sigma_0 m(u)}(v-u)x(v)dv
%\end{equation}
%
$ 
with $u$ a pixel location, $m(u)$ the mask value at $u$ and $\sigma_0$ the maximum isotropic standard deviation of the Gaussian blur kernel $g_{\sigma_0}$, $\sigma_0=10$.
The function $f_c(\cdot)$ is the classification score of the model for the class $c$: the idea is to find a mask that perturbs the original image in a way that the classifier gets maximally confused, rejecting the sample for that class. The second member of Eq.~(\ref{eq:orig}) is a L1-regularizer with strength $\lambda_1$, which guides the optimization to minimally perturb the pixels of the input image. The authors of~\cite{Fong_2017_ICCV} suggested also a total variation (TV) regularizer $\sum_{u \in \Lambda} \left\lVert \nabla m(u) \right\rVert _\beta ^\beta$, in which the sum operates on the $\beta$-normed partial derivatives on $m$, calculated as the difference of the values of two contiguous pixels according to the direction. 

We contribute here by adding a sparsity regularizer $ \sum_{u \in \Lambda} \lvert 1 - m(u) \rvert m(u)$ enforcing sparsity~\cite{tibshirani1996regression} in the values of the mask $m$, making it binary. This regularizer has been designed to start working after a certain number of iterations, so we can get a rough version of the mask before starting to optimize its crisp version, in line with the MacKay's scheduler of~\cite{mackay1995probable}.
The final version of the optimization is thus:
\begin{multline}
\label{eq_sparseReg}
m = \argmin_{m \in \left[ 0, 1 \right]^\Lambda} f_c(\Phi(x;m) ) + \lambda_1 \left\lVert 1 - m \right\rVert_1 + \lambda_2 \sum_{u \in \Lambda} \left\lVert \nabla m(u) \right\rVert _\beta ^\beta + \lambda_3 \sum_{u \in \Lambda} \lvert 1 - m(u) \rvert m(u)
\end{multline}
With $\lambda$s and $\beta$ values set to $\lambda_1 = 0.01$, $\lambda_2 = 0.0001$, $\lambda_3 = 0$ and $\beta = 3$ during the first $300$ iterations. We then modified the parameters to $\lambda_2 = 1$, $\lambda_3 = 2$ for the next $150$ iterations. At the end of the mask extraction stage, each image $x_i$, $i=1...N$ of a given class becomes associated to the corresponding mask $m_i$.

\subsection{Clustering}
\label{method_visualsummarization}

\begin{figure*}
   \includegraphics[width=1.0\linewidth]{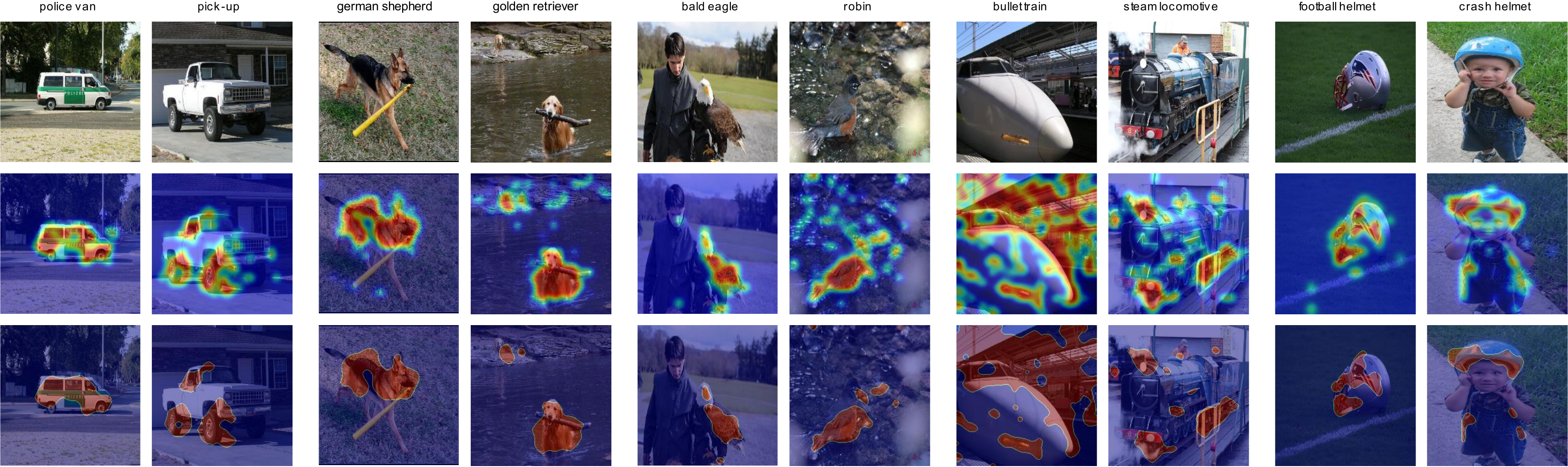}  \vspace{-0.6cm}
    \captionof{figure}{Qualitative analysis of the masks. First row, original image from different Imagenet classes. Second line, heatmaps computed with the method proposed by \cite{Fong_2017_ICCV}. Third line, crisp masks computed with our optimization procedure. Best in colors.} \label{fig:figqualitativemask}
\end{figure*}

Each saliency mask $m_i$ can be analyzed by considering its connected components $\{r_{j}^{(i)}\}_{j=1...J_i}$ called here \emph{regions}. Some of the regions are to be clustered together across multiple images of the same class to form the visual summaries of that class. The idea is that each region represents an articulated visual item composed by \emph{parts}, and a summary is an ensemble of regions exhibiting at least a common part. A graphical sketch of the procedure is shown in Fig.~\ref{fig:figmethodcluster}. 

In our implementation, object proposal technique ~\cite{selectivesearch2013} is employed to extract the parts of the regions. Next, the proposal flow technique~\cite{proposalflow2016} is incorporated to cluster the regions. Indeed, object proposals have been found well-suited for matching, with the proposal flow exploiting local and geometrical constraints to compare structured objects exhibiting sufficiently diverse poses~\cite{proposalflow2016}. 

Our procedure begins by considering the whole images of a class without resorting to the regions, in order to account as much as possible of the context where regions are merged. 
Given a class, all of its $N$ images are processed; from image $x_i$, the set of object proposals $P_i$ is extracted. Next, all of the images are pairwise matched adopting the proposal flow algorithm. Each pair of images $<x_i,x_j>$ will thus produce a $M_i\times M_j$ matrix $Q_{ij}$, with $M_i$ indicating the number of object proposals found in image $x_i$. Each entry of the matrix $Q_{ij}(k,l)$ contains the matching compatibility between the $k$-th and the  $l$-th object proposal of the images  $x_i$ and $x_j$, respectively.

After this step, all the object proposals of all the pairs of images are combined together into a $N_P \times N_P$ matrix $Corr$, where $N_P=\sum_{i=1...N}M_i$ is the total number of object proposals. A given row of $Corr$ will contain the matching score of a particular object proposal with all the remaining object proposals. $Corr$ could be very large but can made easily sparse by thresholding the minimal admissible matching score.  

At this point, we refer to the image regions $\{r_{j}^{(i)}\}$ extracted earlier and select from $Corr$ all of the object proposals that overlap sufficiently with a region (overlap ratio higher than 75\%). In the case of two overlapping proposals, one of them is removed if the ratio between the two areas is less than a certain threshold (2 in this work). The pruning stage leads to the $Corr''$ matrix.

The matrix $Corr''$ is considered as a similarity matrix, and the Affinity Propagation clustering algorithm is applied~\cite{affinitypropagation2007} on top of it. Affinity Propagation requires only one parameter to be set (making parameter selection easier) and it is able to discover the number of clusters by itself. The resulting clusters are ensembles of parts which, thanks to the proposal flow algorithm, should consistently identify a particular portion of an articulated object, thus carrying a clear visual semantics.  Next, post-processing is carried out to prune out unreliable clusters. To this end, Structural Similarity Index (SSIM) \cite{ssim2004} is applied to all the pairs of a cluster, discarding it as inconsistent if the median value of SSIM for that cluster is lower than a threshold based on the global median of SSIM within the whole class (90\% in this work).  This has the purpose of removing obvious mistakes in the clusters, caused by the variety of different poses that the proposal flow has not been able to deal with\footnote{Experimentally we found that in some cases of objects oriented in opposite directions, like cars towards right and left, proposal flow did not work properly providing erroneously high matching scores, as for some complex not rigid objects like animals in drastically different poses.}.   

%% Per il futuro: provare biclustering!
All the parts of a valid cluster are highlighted in red and shown surrounded by the regions they belong to; this eases the human interpretation and provides a summary (see an excerpt in Fig.~\ref{fig:figtitle}).  An explanation is provided for each image class using a  different number of summaries, depending on the number of valid clusters that have been kept.

\section{Experiments}
\label{experiments}
\begin{figure}
\begin{floatrow}
\ffigbox{%
      \includegraphics[width=1.0\linewidth]{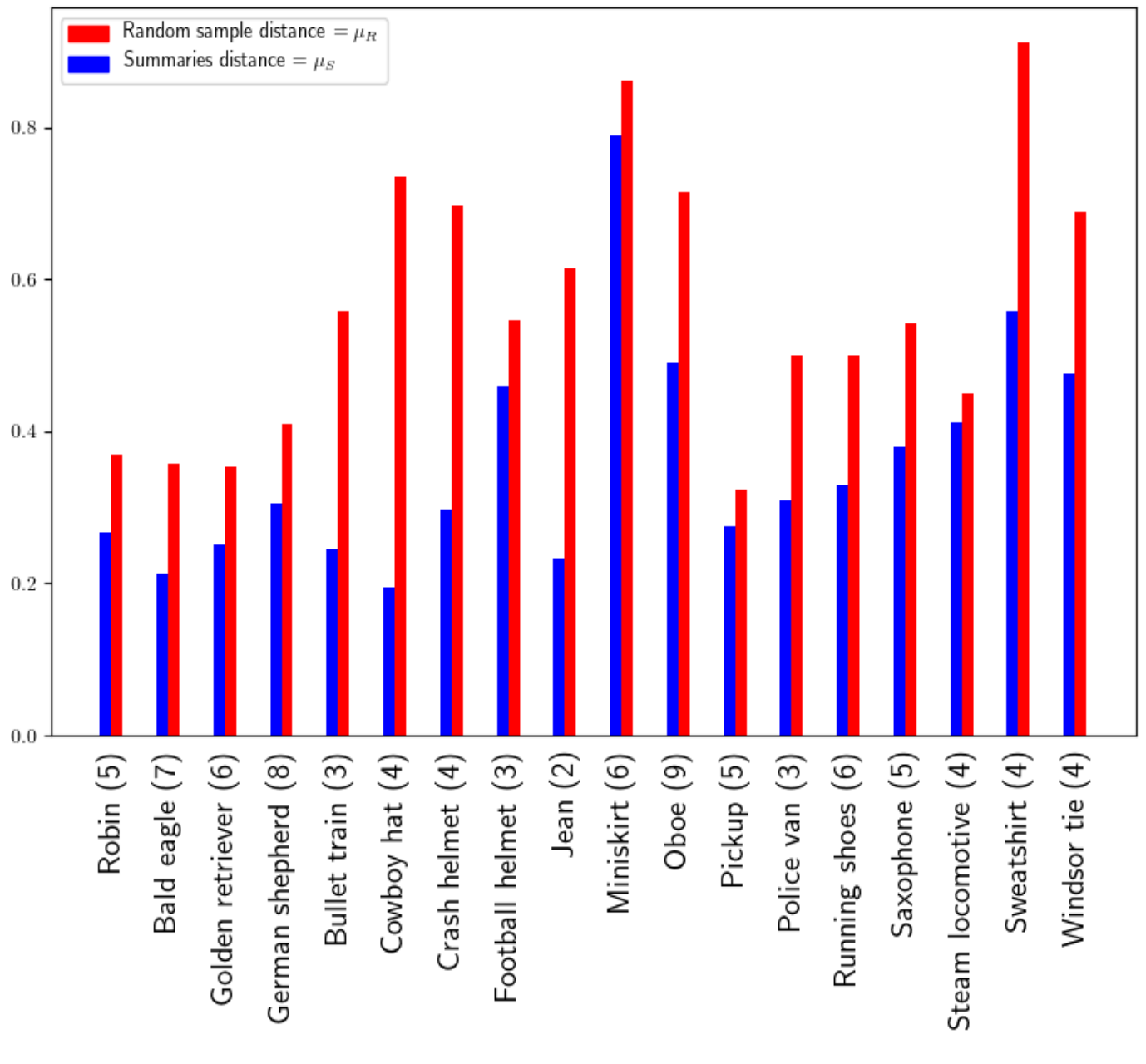} \vspace{0.0cm}
}{%
  \vspace{-0.5cm}\captionof{figure}{Coherency in terms of average Jaccard distance (y-axis) among the tags found with the automatic tagger, within the summaries (blue $=\mu_S$), and within a random sample of the class (red $=\mu_R$). \emph{Lower is better}. The class labels come with the number of summaries found.\label{fig:autotaggerjaccard}} 
}
\capbtabbox{%
	\tiny
\begin{tabular}{|l|l|l|}%{|p{1.8cm}|p{0.7cm}|p{5.3cm}|}%
\hline
Class Name & $\mu_U$ & Most Proposed Tag per Summary\\
\hline
\hline
\rowcolor[HTML]{C0C0C0}  
Robin & 0.12 & Head, Body, Legs, Wings, Tail                                                                                 \\ \hline
\rowcolor[HTML]{C0C0C0}  
Bald eagle & 0.23 & \begin{tabular}[c]{@{}l@{}} Head, Neck border, Eye, Beak 
Face, Wing \end{tabular} 	                                                                                       \\ \hline
\rowcolor[HTML]{EFEFEF}
Golden retriever & 0.31 & \begin{tabular}[c]{@{}l@{}}  Nose, Eye, Ear, Mouth, Face, Legs, Head \end{tabular}                                                                                             \\ \hline
\rowcolor[HTML]{EFEFEF} 
German shepherd & 0.22 & \begin{tabular}[c]{@{}l@{}}  Eye, Leg, Neck, Body, Ear, Nose, Face, Feather  \end{tabular}                                                                                 \\ \hline
\rowcolor[HTML]{C0C0C0} 
Bullet train & 0.38 & \begin{tabular}[c]{@{}l@{}} Front train, Front glass, Train, \\ Rails, Lights, Train body  \end{tabular}                                                                       \\ \hline

\rowcolor[HTML]{C0C0C0}
Steam locomotive & 0.56 & \begin{tabular}[c]{@{}l@{}}  Chimney, Front train, Wheels, \\ Engine, Side, Window \end{tabular}                                                                                        \\ \hline
\rowcolor[HTML]{EFEFEF}
Pick-up & 0.19 & \begin{tabular}[c]{@{}l@{}} Mudguard, Step bumpers, Side window,\\ Windshield,Back, Wheel \end{tabular}                        \\ \hline

\rowcolor[HTML]{EFEFEF}
Police van & 0.17 & \begin{tabular}[c]{@{}l@{}} Wheel, Police flag, Side window, Light, \\ Rear window, Vehicle, Capote, \\ Bumpers, Mudguard \end{tabular} \\ \hline 

\rowcolor[HTML]{C0C0C0}
Oboe & 0.01 & Body, Buttons                                                                                                  \\ \hline
\rowcolor[HTML]{C0C0C0}
Saxophone & 0.68 &Body, Buttons, Bell                                                                                         \\ \hline
\rowcolor[HTML]{EFEFEF}
Crash helmet & 0.36 & Base, Side, Front, Logo                                                                                             \\ \hline
\rowcolor[HTML]{EFEFEF}
Football helmet & 0.48 & Front grids, Logo, Side, People                                                                                             \\ \hline
\rowcolor[HTML]{C0C0C0}
Jeans & 0.01 &Crotch, Pocket, Legs, Waistband                                                                            \\ \hline
\rowcolor[HTML]{C0C0C0}
Miniskirt & 0.12 &Face, Waistband, Leg, Head                                                                                \\ \hline
\rowcolor[HTML]{C0C0C0}
Cowboy hat & 0.32 & Ear, Face, Chin                                                                                  \\ \hline
\rowcolor[HTML]{C0C0C0}
Windsor tie & 0.13 & Pattern, Knot, Collar, Neck                                                                              \\ \hline
\rowcolor[HTML]{C0C0C0}
Sweatshirt & 0.31 & \begin{tabular}[c]{@{}l@{}}Hoodie, Face, Arm, Laces, Wrinkles, Neck \end{tabular}                                                                        \\ \hline
\rowcolor[HTML]{C0C0C0}
Running shoes & 0.38 & Laces, Logo, Shoe side                                                                                     \\ \hline
\end{tabular} \label{table_classes} \vspace{+0.4cm}
}{%
  \vspace{-0.0cm}\caption{Classes from ImageNet, coherency of the summaries in terms of average Jaccard distance  ($=\mu_U$) among the tags found with the user study and the set of tags collected during the user study with our approach.\label{tab:table_classes} %The textual tags are consistent with the visual summaries given in Fig. \ref{fig:figtitle} and saliency masks in Fig. \ref{fig:figqualitativemask}.
}
}
\end{floatrow}
\end{figure}

For our experiments, we focus on 18 classes of Imagenet. These classes are selected considering the constraint of being adjacent in a \emph{dense}~\cite{deng2010does} semantic space. In Table~\ref{tab:table_classes}, adjacent classes are in subsequent rows with same background color. This constraint, brings together those classes that are adjacent to each other which provides the possibility of comparing \emph{similar} classes along different experiments. 

The set of experiments to validate our proposal is organized as follows: Sec.~\ref{Sec:exp-Masks} is dedicated to show the superiority of our proposed crisp mask w.r.t. the original smooth mask \cite{Fong_2017_ICCV} in terms of conciseness and expressiveness, providing higher classification drop. Sec.~\ref{Sec:exp-Summary} is focused on the semantics of the summaries, showing that automatic taggers as well as humans, individuate a precise type of parts for each summary. Sec.~\ref{Sec:exp-Num} shows that the number of summaries is proportional to the classification ability of a deep architecture: the higher the number of classes the higher the classification accuracy.  In Sec.~\ref{Sec:exp-Fine} it is showed that summaries can be used to specialize the classifier on the visual summaries and improve the classification results.

\subsection{Masks analysis}\label{Sec:exp-Masks}

In this experiment the masks obtained by our approach are compared with those of the smooth mask a.k.a. IEBB \cite{Fong_2017_ICCV} method employing the protocol as proposed by the authors. Given an image, the classification confidence associated to it w.r.t the ground truth class is measured. In the case of a deep network, the classification confidence for the $i$-th object class is the softmax output in the $i$-th entry. Afterwards, the image $x$ is blurred as explained in Sec.~\ref{method_saliency} by using the corresponding mask $m$ (either the one produced by our proposed approach or the one produced by the IEBB approach). The classification score is then re-computed after perturbation and the difference w.r.t. the score for the original image is computed. The average classification drop of a method is computed as the average score drop over the entire test set. We compare our proposal solely with IEBB, which is shown to be the state-of-the-art~\cite{Fong_2017_ICCV}. In addition, we compare with IEBB \emph{thresh}, in which the smooth mask generated by IEBB is made crisp by a thresholding operation over the mask intensities. 
On each image the threshold is independently set to make the mask as big as the one produced by our proposed technique to ensure a fair comparison. The third column of Table~\ref{tab:quantitative} shows the classification loss of the two approaches. Notably, we succeed in improving the results, closely reaching the saturation. Interestingly, with IEBB \emph{thresh}, the overall performance diminishes, with higher variance.

In Fig.~\ref{fig:figqualitativemask}, examples of the obtained masks using our approach and IEBB are shown. From our observations, the sparse optimization producing mask which are similar to the IEBB one. In fact, IEBB finds masks which cause a nearly complete loss. Nonetheless, our improvement gives the same importance to all of the pixels which leads to a higher classification drop, while facilitating the clustering step and consequently the final human interpretation of the summaries. 

\subsection{Analysis of the summaries}\label{Sec:exp-Summary}
In this section of the experiments, we make use of an automated tagger~\cite{imagga} to show whether each summary individuates a visual semantic. For each object class, the $n_i$ images of each single summary $S_i$, $i=1,...,K$ are tagged, providing $n_i$ lists of textual tags (only nouns are allowed). For convenience, the tagger is constrained to provide only 8 tag for each image. This procedure is repeated on $K$ sets $R_i$, $i=1,...,K$ of $c_i$ random images taken from that class.

\begin{figure}
\begin{floatrow}
\capbtabbox{%
\small
	\begin{tabular}{|l|c|l|}
  \hline 
  Method & Ref. & \%Drop (Var) \\
  \hline
  \hline
  IEBB & ICCV17\cite{Fong_2017_ICCV} & 99.738365 (8.13e-4) \\
  IEBB \emph{thresh.} & ICCV17\cite{Fong_2017_ICCV} & 97.703865 (5.758e-3) \\
  \textbf{Ours} & & \textbf{99.964912}($<10$e-6) \\
  \hline
  \end{tabular}
}{%
\caption{Mask analysis results. \label{tab:quantitative}}
}
\capbtabbox{%
\small
	\begin{tabular}{|l|l|l|}
\hline 
  Model & Summaries & Acc. \\
  \hline
  \hline
  AlexNet & 5 & 57.1\% \\ \hline
  VGG16& 5.5 & 72.4\% \\ \hline
  GoogleNet& 6 & 74.5\% \\ \hline
  Resnet50& 6.33 & 76.2 \% \\ \hline
\end{tabular}
\vspace{-0.2cm}
}{%
  \caption{Average number of summaries for each different architecture and top-1 accuracy.\label{table_architectures}}
}
\end{floatrow}
\end{figure}

After tagging, the set of all the given tags is used to extract a one-hot vector for each image. The entry of the vector is 1 if a particular tag is given, and 0 otherwise. Synonyms tags were fused together by checking synsets of WordNet. This results to a vector of an average length of 28 entries. At this point, the $n_i$ tag vectors of the summary $S_i$ are pairwise compared with the Jaccard distance, and the average intra-summary distance is computed. This is computed for each summary, and the $K$ average intra-summary distances are further averaged, obtaining the summary distance $\mu_S$. This process is repeated for each class. In the same way, we compute the average distance obtained with the random image subsets $R_i$, getting a $\mu_R$ for each class. Results are shown in Fig. \ref{fig:autotaggerjaccard}. As it can be seen, on average images belonging to the same summary are closer in semantic content (i.e. lower Jaccard distance) than random images of the same class.

%It is worth to mention that, the previous analysis is performed on the entire image content, while our summaries are characterized by regions of images (the crisp mask) enriched by a bounding box addressing the part. 
%It is worth to mention, that the previous analysis is performed on a much noisier data, since we are not able to focus its attention on an arbitrary part of the image. We expect that the semantics that can be extracted from the summaries is much fine grained. 
Since the automated tagger could only work on the entire image, we expect to have much finer grained results by focusing on the parts highlighted by the summaries.
To this end we organize a user study, with the goal of giving a precise name to each of the summary, by considering the parts highlighted within. We hire a total of 50 people (35 male, 15 female subjects) with an the average age of 33 (std:8.4). Each of the users was asked to give a set of (noun) tags to each summary, by considering the entire set of regions and parts contained within. 
Next we check the inter/rater reliability among users toward the same summary by computing the average pairwise Jaccard distance among the obtained sets of tag. The distances over the different summaries are averaged, thus obtaining for each class $\mu_U$ which is a measure of the agreement between users expressed as the average . To name each summary, we select the tag more used among the users. Table~\ref{tab:table_classes} report on the right these tags (one for each summary), together with the $\mu_U$ value. Interesting observations can be assessed: in some cases, the $\mu_U$ values are very small, but at the same time many tags are definitely more specific than those provided by the automatic tagger, indicating that the summaries individuate finer grained visual semantics that users have captured. Then, adjacent classes exhibit many common visual summaries (\emph{german shepard, golden retriever}).

\begin{figure*}
   \includegraphics[width=1.0\linewidth]{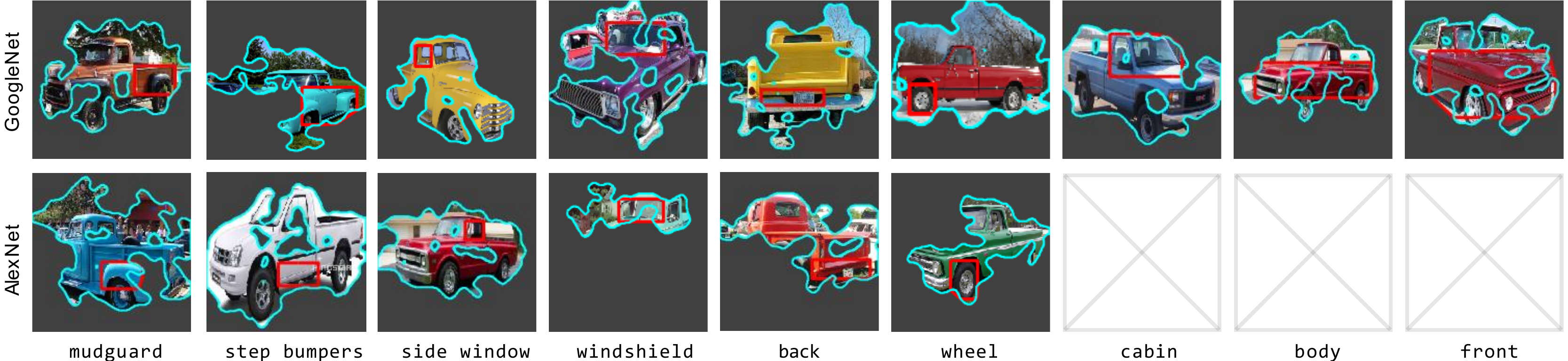} \vspace{-0.5cm}
    \captionof{figure}{Motivating the superiority of GoogleNet against AlexNet. focusing on the \emph{pick-up} class, our approach finds 9 summaries for the former architecture, 6 for the latter, showing that GoogleNet is capable of capturing more semantics. Best seen in color.}\label{fig:figclustcomparison} 
\end{figure*}

\subsection{Number of summaries and classification accuracy}
\label{Sec:exp-Num}

Another interesting question to be answered is whether the number of summaries has a role in the general classification skill of a network. To this end, we analyze four famous architectures as, AlexNet \cite{krizhevsky2012imagenet}, VGG \cite{simonyan2014very}, GoogleNet \cite{szegedy2015going}, and ResNet \cite{he2016deep}. For each of these architectures, the average number of summaries over the 18 chosen classes for the analysis is computed. This value is later compared with the average classification ability of each architecture in terms of accuracy over ImageNet validation dataset. The comparison results are shown in Table~\ref{table_architectures}. Notably, from AlexNet to ResNet, as the classification accuracy rate increases, the number of summaries also rises. From this observation, we can conclude that the network classification ability is related to the the number of discriminant patterns that the network is able to recognize. This has been shown qualitatively in Fig.~\ref{fig:figclustcomparison}. We obtained similar observations with other classes and other architectures.

\begin{figure*}
\includegraphics[width=1.0\linewidth]{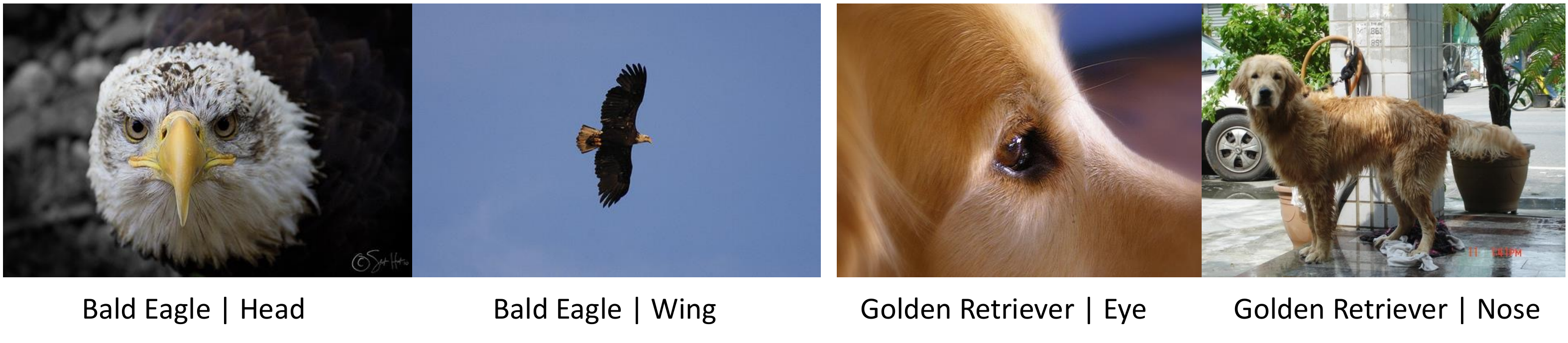} \vspace{-0.7cm}
\captionof{figure}{Examples of images two classes that were misclassified by the AlexNet but correctly classified by specializing the classification using SVMs trained on the summaries. The labels below are the class names and the tags associated with the summary that contributed the most to correcting the classification of each image.}\label{fig:fig_nnwrong_svmyes} 
\end{figure*}

\subsection{Specializing classification with the summaries}\label{Sec:exp-Fine}
The proposed idea in this section is to improve the classification results using the images belonging to the summaries. Due to the low number of images per summary (average of 32.25), we propose to employ a linear SVM per summary instead of explicitly fine-tuning the network itself. Positive examples to train each SVM are the images belonging to that summary, and negative examples are images from other classes or from other summaries within the same class. The features used for classification are extracted from the first fully connected layer of the network. Given an image to classify, it is evaluated by all of the previously trained SVMs. The class scores vector is then obtained by selecting the highest score among the SVMs for each class. The obtained scores are used to improve the classification accuracy for a desired class by means of a convex weighted sum between the neural network classification softmax vectors and the resulting SVM class scores (normalized to sum to unity). Our experiments show that employing this approach, primarily designed to improve the classification of all the 18 classes chosen for the experiments on the AlexNet architecture, the overall classification accuracy score over all the 1000 ImageNet classes increases by 1.08\% on the ImageNet validation set. Some examples of images that are classified correctly thanks to this boosting technique can be seen in Fig.~\ref{fig:fig_nnwrong_svmyes}.

\section{Conclusion}
\label{conclusions}
Our approach is the first visualization system which considers multiple images at the same time, generalizing about the visual semantic entities captured by a deep network. Contrarily to the standard visualization tools, advantages of our proposed approach can be measured quantitatively, the most important of them is that of improving the original network by training additional classifiers specialized on recognizing the visual summaries. The future perspective is to inject the analysis of the summary in the early training of the deep network, and not only as a post processing boosting procedure.

\bibliography{bibliography}
\end{document}